\RequirePackage{snapshot}
\documentclass[num-refs]{sn-jnl}
\usepackage{xcolor}
\usepackage{graphicx}
\usepackage{hyperref}

\usepackage{setspace}

\usepackage{lineno}

\usepackage{amsmath}

\usepackage[utf8]{inputenc} 

\usepackage[numbers]{natbib}

\newcommand{\todo}[1]{}

\newcommand{\todoRev}[2]{\todo{(from rev. #1): #2}}
\newcommand{\ans}[1]{RESOLVED at line \ref{#1}}
\renewcommand{\linelabel}[1]{\label{#1}}

\title{A Generative-AI Modeling Framework for Explainable Decision Support in Complex Geosteering Scenarios}

\author*[1]{\fnm{Sergey} \sur{Alyaev}}\email{saly@norceresearch.no}
\author[1]{\fnm{Kristian} \sur{Fossum}}
\author[1,2]{\fnm{Hibat Errahmen} \sur{Djecta}}
\author[1]{\fnm{Jan} \sur{Tveranger}}
\author[3]{\fnm{Ahmed H.} \sur{Elsheikh}}

\affil[1]{\orgname{NORCE Norwegian Research Centre AS}, \orgaddress{\city{Bergen}, \country{Norway}}}
\affil[2]{\orgname{University of Stavanger}, \orgaddress{\city{Stavanger}, \country{Norway}}}
\affil[3]{\orgname{Heriot-Watt University}, \orgaddress{\city{Edinburgh}, \country{United Kingdom}}}

\abstract{
The real-time process of directional changes while drilling, known as geosteering, is crucial for hydrocarbon extraction and emerging directional drilling applications such as geothermal energy, civil infrastructure, and CO2 storage. The geo-energy industry seeks an automatic geosteering workflow that continually updates subsurface uncertainties and captures the latest geological understanding, informed by real-time observations. 

We propose a real-time, AI-driven geosteering workflow that integrates Generative Adversarial Networks (GANs) for geological parameterization, ensemble methods for model updating, and global discrete dynamic programming (DDP) optimization for complex decision-making during directional drilling operations. Our framework relies on offline training of a GAN model to reproduce relevant geology realizations and a Forward Neural Network (FNN) to model the response of Logging-While-Drilling (LWD) tools for a given geomodel.

This paper introduces a first-of-its-kind workflow that progressively reduces GAN-geomodel uncertainty around and ahead of the drilling bit and adjusts the well plan accordingly. The workflow automatically integrates real-time around-bit LWD, which, through learned geological correlations, reduces uncertainty in predicted geology ahead of drilling.
A DDP-based decision support system leverages probabilistic look-ahead predictions to suggest better steering strategies. 
We test the workflow prototype on a small yet challenging low-net-to-gross drilling scenario with several possible targets. The results show that the workflow produces meaningful steering recommendations and, through its probabilistic updates, automatically maps formation boundaries along the drilled well. }

\begin{document}

\maketitle
% \modulolinenumbers[5] \linenumbers    

\section{Introduction}

Dynamic real-time adjustment of well trajectory, or geosteering, is established in hydrocarbon drilling operations to maximize efficiency and safety \citep{hermanrud2019future}. Traditionally, geosteering has depended heavily on the manual interpretation of real-time Logging-While-Drilling (LWD) data, a challenging task under the pressure of rapid drilling operations and complex geological uncertainties \citep{cheraghi2022can, rassenfoss2022study}. The application of geosteering extends beyond hydrocarbon wells to geothermal \citep{ungemach2021real}, civil infrastructure, and emerging domains like CO2 storage \citep{miotti2022machine}, further underscoring its importance.

Current methodologies developed for well-planning incorporate Ensemble-based Closed-Loop Reservoir Management (EnCLRM), which has set a new standard for probabilistic estimate of subsurface uncertainties \citep{hanea2016decision, skjervheim2012integrated, abadpour2018integrated} and informed decision support \citep{hanea2019robust}. Despite its effectiveness, implementing EnCLRM in real-time scenarios is limited by the computational demands of its complex modeling sequences. Within these sequences, the geomodeling that involves depth conversion, structural modeling, faults, facies modeling on a grid, and property modeling \citep{skjervheim2012integrated} is among the hardest to automate and adapt for real-time operations.

Recent technological innovations have introduced generative machine learning, including variational auto-encoders \citep{canchumuni2019towards}, Generative Adversarial Networks (GAN) \citep{chan2017parametrization}, and, lately, latent diffusion models \cite{lee2025latent,di2025latent} as powerful tools for geomodeling. 
We will refer to them as GANs despite meaning a larger class of models. These models, capable of rapidly producing realistic geomodel realizations from a reduced set of parameters, provide a promising approach to overcoming traditional limitations. \citet{alyaev2021probabilistic} and \citet{fossum2022verification} demonstrated that GANs combined with ensemble methods enable quick updates to geological models in response to acquired drilling data. Thus, GAN-based workflows can potentially transform geosteering into a more quantifiable and precise process. 

While in field operations, decisions remain the responsibility of a geosteering team \citep{hermanrud2019future, rassenfoss2022study}, academic studies have advanced decision optimization methods to enhance efficiency and consistency. Initial approaches, such as greedy optimization, were employed by \citet{chen2015optimization} and \citet{kullawan2016value}, focusing on immediate gains at each decision stage, but often resulted in sub-optimal outcomes due to the lack of future consideration \citep{kullawan2018sequential}. \citet{pavlov2024geosteering} expanded on greedy optimization by integrating real-time formation evaluation data and differential evolution algorithms to optimize drilling trajectories, showing improvements over traditional methods.

To address the limitations of greedy optimization, \citet{kullawan2018sequential} introduced Approximate Dynamic Programming (ADP) for geosteering, which incorporates future decisions and learning for proactive, globally optimized decision-making, superseding the greedy approaches in many scenarios. Despite its effectiveness, ADP's high computational demands and scenario-specific design pose challenges for real-time applications. 
\citet{alyaev2019decision} proposed a version of ADP, termed 'naive-optimistic' \citep{alyaev2018you}, that adapted the method to a general updatable ensemble representation of uncertainty, albeit with a loss of proven dynamic-programming (DP) optimality. 
\linelabel{naiveoptimistic}
Nevertheless, it outperformed most experts in a synthetic experiment \citep{alyaev2021interactive}. Outside of geosteering, for sequential well location selection to avoid subsurface caves, \citet{kanfar2024well} explored the application of GANs combined with the Fast Informed Bound (FIB) algorithm, which utilizes elements of ADP to compute upper bounds on the value function.

Recently, reinforcement learning (RL) has emerged as a promising alternative to ADP methods, offering flexibility and computational efficiency once trained.
RL trains an agent through iterative learning from environmental interactions, thereby developing optimal decision-making strategies suited for real-time scenarios. 
\citet{kristoffersen2021automatic} presented the use of the 'NEAT' evolutionary RL algorithm to incorporate the effects of geosteering during well planning, demonstrating another aspect of decision optimization.
\citet{muhammad2023optimal} demonstrated that Deep Q-Network (DQN) algorithm achieves comparable and sometimes better results than ADP in geosteering optimization. 
\citet{muhammad2025high} combined DQN with particle filter, that in several experiments outperformed all the experts
in the close-to-realistic ROGII Geosteering World Cup 2021, unconventional stage
\citep{muhammad2025geosteering}.

This paper introduces a holistic geosteering approach that integrates an updatable ensemble of GANs within a probabilistic framework for dynamic well placement. 
The workflow is modular, allowing for the combination of different generative geomodels and various measurements, with diverse data assimilation and decision support algorithms, making it easily adaptable to other geosteering scenarios.
Our initial implementation utilizes real-time ultra-deep electromagnetic (UDAR) data and a robust Decision Support System (DSS), combining the Ensemble Kalman filter (EnKF), GAN-based geomodels, and the adapted ’naive-optimistic’ dynamic programming.  The workflow enhances the precision of real-time subsurface modeling and decision-making by automating geosteering through a closed loop of sequential decisions and data acquisition. It moves towards an ’autopilot’ capability for complex geosteering challenges. 

This paper extends a conference paper \cite{alyaev2024distinguish} and reuses parts of the methodological description and the text. 
The rest of the manuscript is organized as follows:
Section~\ref{sec:method} details the proposed workflow, including the multi-step forward mapping sequence, the EnKF data-assimilation procedure, and the dynamic-programming-based DSS under uncertainty. 
Section~\ref{sec:results} presents the synthetic reference case: prior construction, objective and constraint formulation, data-assimilation setup, sequential decision recommendations, and a discussion of performance and limitations. 
Section~\ref{sec:concl} summarizes the main contributions, outlines observed artifacts and limitations, and highlights avenues for future developments.

\section{Method}
\label{sec:method}
The proposed workflow consists of two parts: the offline (pre-job) phase and the online phase. The online phase relies on selecting a geological prior and is enabled by offline training: we train a GAN model to reproduce relevant geology realizations and a Forward Neural Network (FNN) to model LWD tools' response for a given geomodel. The offline neural-network training phase (similar to \cite{alyaev2021probabilistic}) is described in the appendix, Section \ref{sec:appendixTraining}.

\begin{figure}
    \centering
    \large
    (a)\\
    \includegraphics[width=\textwidth]{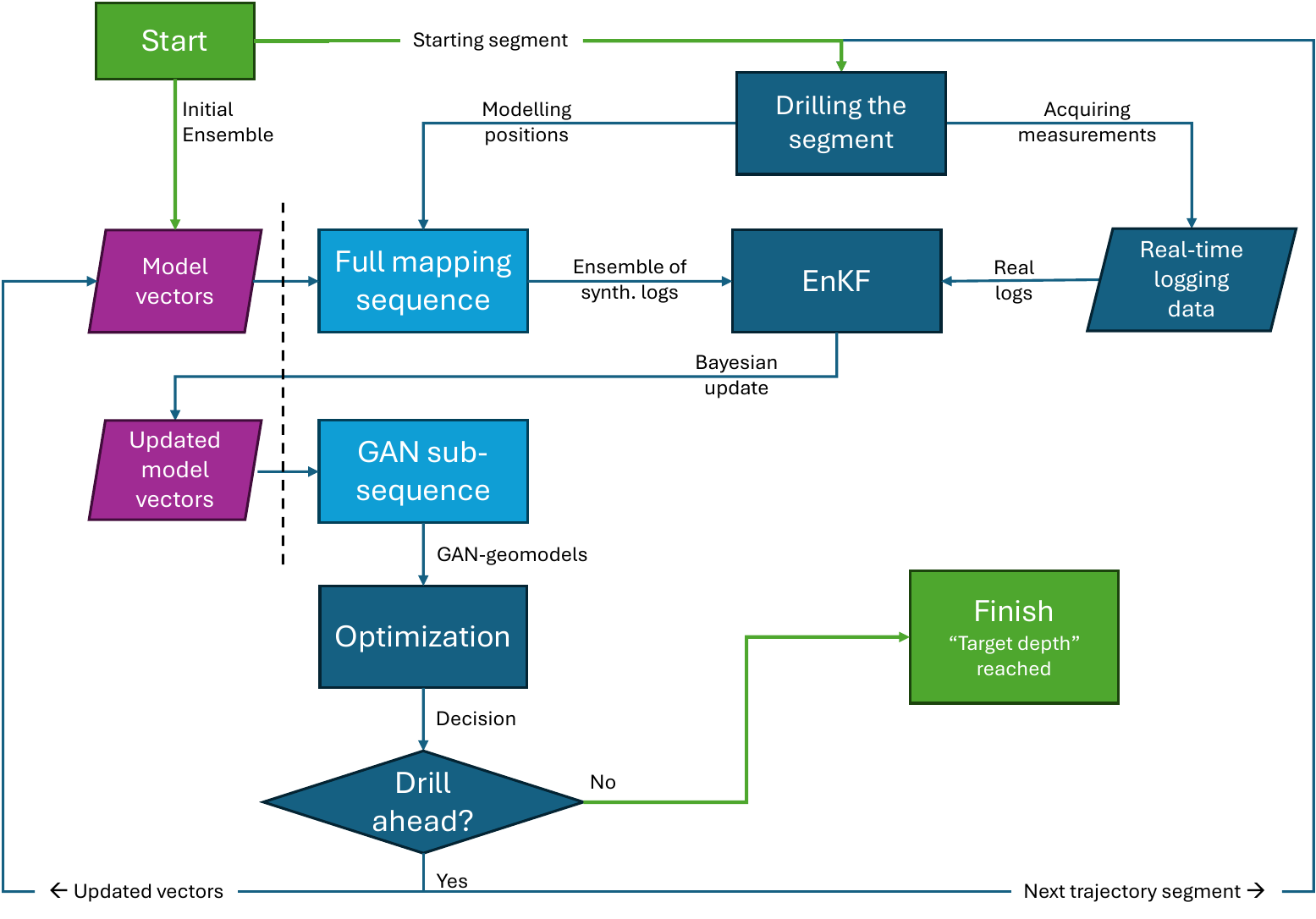}\\
    (b)\\
    \includegraphics[width=0.8\textwidth]{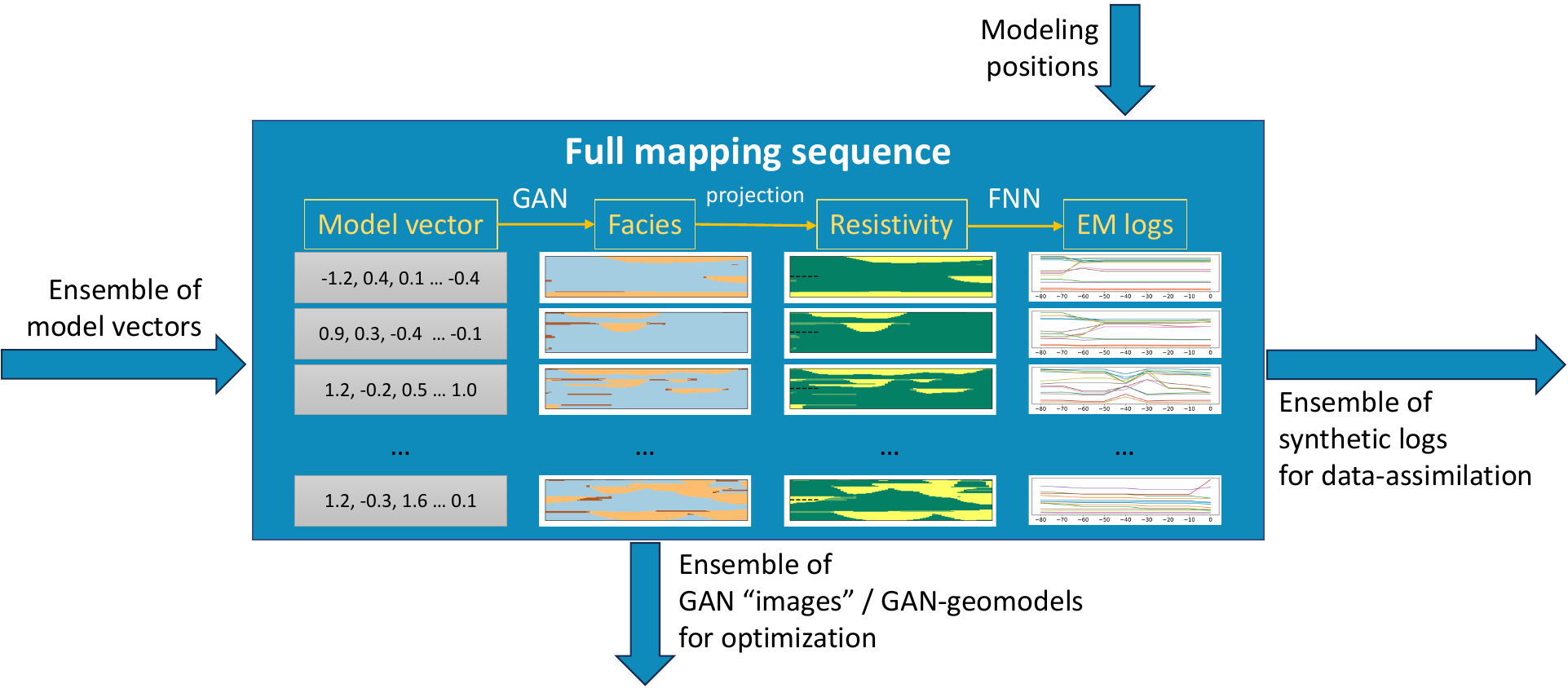}
    \caption{Overview of the proposed workflow: (a) the online phase, and (b) a detailed schematic of the "full mapping sequence" highlighting its inputs and outputs.}
    \label{fig:workflow}
\end{figure}

\todoRev{3}{SOLVED \ref{gaussianity}: refer to appendix. 
Are the 60 variables considered independent? If yes, during the offline training, how is this assumption honored? Is the assumption of Guassianity valid? If so, why?}

Figure \ref{fig:workflow}(a) depicts the online phase of the workflow. It begins with the initialization of an ensemble of model vectors, encoding various geological and petrophysical properties necessary for subsequent simulation and optimization tasks into a 60-dimensional multivariate normal distribution.
The AI-driven sequence that maps the model vectors to these quantities of interest is central to the workflow; see Figure \ref{fig:workflow}(b).

\todoRev{3}{RESOLVED! Facies maps are non-Gaussian and discrete, while the GAN models I assume are continuous for differentiability needed for training. So how are the discrete realizations are created using a continuous model?}
The mapping sequence (similar to the one used in \cite{alyaev2021probabilistic,fossum2022verification, alyaev2024distinguish}) starts with the model vectors undergoing a GAN-generator multi-step deconvolutional mappings from independently-distributed latent variables to an spatial geological image where three channels of each pixel are interpreted as the probability of each facies at that cell. \linelabel{gancontinuity}
Picking the most probable facies  (minimal rounding in well-trained GAN) produces local geomodel representations, which are required later for optimization.
\todoRev{3}{RESOLVED! What is the benefit of this multi-step approach, because a simpler approach maybe to create a single ML model that directly generates the EM logs as a function of the 60 input features?}
Subsequently, columns of these images (which we call GAN-geomodels) are passed as input to the FNN model which acts as a fast proxy for simulating the modern UDAR measurements. Unlike the previous implementations of GAN-based mapping sequence which used the previous generation extra-deep EM logs \cite{alyaev2021probabilistic,fossum2022verification, alyaev2024distinguish}, in this paper we introduce the full suite of UDAR measurements for which we are also able to provide an open-source proxy model (Section \ref{sec:appendixTraining}). \linelabel{newUDARmodel}
Its output is an ensemble of expected well-log responses based on the geomodels.

Meanwhile, real well logs are acquired during the drilling process in the same positions. These logs offer actual measurements of the subsurface properties and are crucial for reducing uncertainty in the model predictions.

\todoRev{3}{RESOLVED! == link to the book \cite{elsheikh_gan_2026_inpress}? == It is well known that the EnKF is only valid for gaussian random fields, in this case the facies maps are non-Gaussian.
Since the forward model is very fast, why not use inverse modeling techniques like MCMC which are more suited to non-Gaussian
processes?}
The EnKF computes the statistical misfit between the ensemble of predicted measurements and the actual well logs, see Figure \ref{fig:workflow}(a). 
This misfit forms the basis for a Bayesian update of the model vectors, refining them to better match the observed data. According to a previous study, the fast and effective EnKF's linearized update combined with well-trained (unentangled and smooth) GAN \cite{elsheikh_gan_2026_inpress} is sufficient to capture the main modes of uncertainty \cite{fossum2022verification}. \label{enkfWithGan}

The updated model vectors are again converted into GAN-geomodels using the GAN sub-sequence. The GAN-geomodel ensemble now incorporates the latest subsurface information with reduced uncertainty in predicted geology ahead of the drill bit. 

The GAN-geomodel ensemble is input into the decision support system, 
which combines optimization and visual methods to assist in making informed decisions about the remaining drilling trajectory. A dynamic programming optimization method recommends the drilling-trajectory next segment or stopping to maximize the operational objective. 
At the same time, the visualization provides explainability and helps the operational team understand the subsurface structure, the uncertainties, and the DSS recommendations.

The process is sequential, continuously updating and refining the model vectors and the corresponding GAN-geomodels as new real-time data becomes available during drilling. At each step, the team supported by the DSS makes a decision on whether to continue drilling based on the reached and potential targets. If the drilling continues, the workflow loops back, acquiring more real-time logging data and updating the models accordingly. The workflow concludes when the "target depth" is reached, resulting in an optimized trajectory and accurate geological model around the drilled well.

\subsection{Full mapping sequence}

The full mapping sequence of the workflow (Figure \ref{fig:workflow}(b)) consists of several steps enabled by offline training of several neural networks (NNs):

\begin{itemize}
    \item A GAN generator maps model vectors to facies images;
    \item A petrophysical model maps images to resistivity profiles;
    \item An FNN maps resistivities to EM logs along the drilled trajectory.
\end{itemize}
\todoRev{3}{RESOLVED! Some more details of the GAN and the FNN setup and training results as this is the most important part of the paper even though
it has been presented before. } 
The details of the training of neural-network components are provided in the appendix, Section \ref{sec:appendixTraining}.
Splitting training into several explicit networks, where an intermediate network is trained on a more data-rich dataset, can also improve mapping accuracy. 
Adding the direct "skip-connections" from model vectors to the EM logs, and training the FNN with auxiliary-head weighted loss may improve training \cite{lee2015deeply}, but would reduce modularity of the workflow. \linelabel{multistepMapping}
\todoRev{3}{RESOLVED! What is the benefit of this multi-step approach, because a simpler approach maybe to create a single ML model that directly generates the EM logs as a function of the 60 input features?}

The first stage begins with model vectors, which are 60-dimensional in our case. The vector parametrization is learned during the adversarial training of the GAN, which consists of two NNs: a generator and a discriminator, both with deep convolutional architecture. During training, the generator converts vectors selected randomly from a multivariate normal distribution into synthetic facies images, trying to resemble real geological models from the training data. Simultaneously, the discriminator learns to distinguish between real and generated images. 
In this study, the output images have a fixed size of 64x64 cells with three channels, each corresponding to a probability of each facies (channel, crevasse, and floodplain shale).  
This offline training process ensures that the generated facies images are geologically plausible, the corresponding model vectors are distributed as multivariate normal Gaussian, and the mapping between them maps small changes of vectors into small changes in resulting geologies. The full training details are provided in \citet{alyaev2021probabilistic}, and the 3D fluvial geomodel dataset generated using geological software is described in the Appendix.

Following the GAN-generated facies images (GAN-geomodels for short), the next stage involves projecting these images to resistivity models.
In this study, same as \citet{alyaev2021probabilistic}, we employ a petrophysical model that maps each facies to a constant resistivity value, specifically 4.0 ohm-m for floodplain shale, 171.0 ohm-m for oil-saturated channel sand, and 55.0 ohm-m for partially saturated crevasse splay sand. 
This projection captures the electrical properties of the subsurface formations, translating the geological features into resistivity models that can be used for further analysis.

In the final stage, these resistivity models are used to generate synthetic EM-log responses along the trajectory. For each logging position (one per vertical column), we compute the distances to boundaries between facies from the virtual location of the measuring tool (modeling position, see Figure \ref{fig:workflow}(b)). The vertical distances and the layer resistivities are fed into a pre-trained FNN that produces electromagnetic (EM) logs. This study uses the pre-trained forward model FNN described in \citet{alyaev2021modeling} that reproduces the full suite of extra-deep azimuthal resistivity (EDAR) EM logs. In practice, the FNN must be trained for the relevant geological setting \citep{rammay2024strategic}. The EDAR FNN model has up-down sensitivity for up to six boundaries, potentially covering the full GAN-geomodel thickness of 32 meters from any measurement position. 
Thus, EDAR EM logs contain information about the subsurface's electrical properties, which is essential for updating the geomodel and accurate geosteering.

The mapping sequence serves two key purposes: providing geological predictions around and ahead of the bit and transforming these predictions into relevant measurements. The modularity of the workflow allows for flexible adjustments in both areas. The mapping to local geological models - such as facies images - captures critical subsurface configurations and uncertainties, which are required for optimization. At the same time, the mapping to relevant real-time LWD measurements - such as EM logs - informs the model updates under uncertainty. This dual-purpose structure ensures that the workflow remains adaptable. It allows each component to evolve independently as new data or complexities are introduced, making it an effective tool for geosteering and real-time decision-making.

\subsection{EnKF for data assimilation and model updates}

The full mapping sequence, as described above, maps a Gaussian model vector into a synthetic EM-log. Given measurements of real-time logging data, one can then update the model vectors and implicitly the GAN-geomodels by solving the statistical inverse problem. 
Encouraged by the results obtained in~\citep{fossum2022verification}, we employ ensemble-based data assimilation to solve the inverse problem. 
Adapting to the sequential nature of the DSS, we choose the EnKF for data assimilation already demonstrated for other geosteering tasks in \citet{chen2015optimization} and \citet{alyaev2019decision}.

The EnKF updates the system states sequentially using Bayesian principles each time new observations are received. In a geosteering operation, LWD measurements are acquired and transmitted to the surface as the well is drilled, making the EnKF's sequential nature well-suited for this problem. In this context, the system state corresponds to the poorly known earth model, and updates of the system state correspond to solving the inverse problem~\citep{Iglesias2013}.

Considering the model and observations as stochastic variables, we use Bayes' theorem to solve for the conditional distribution up to a normalizing constant
\begin{equation}
    p\left(m|d\right) \propto p\left(d|m\right)p\left(m\right).
\end{equation}
Here, $p\left(d|m\right)$ is the likelihood and $p\left(m\right)$ is the prior distribution of the model. In the EnKF, Bayes' theorem is used recursively to update the model at each observation point by using the last posterior model as the current prior $p\left(m_i\right) =  p\left(m_{i-1}|d_{i-1}\right)$. In the following, we will omit the time index, $i$, and only focus on updates at a single measurement point.

Assuming that all variables have Gaussian distributions and that there is a linear relationship between the earth model and the EM-response, one can derive an analytical solution for the mean and covariance of the conditional distribution~\citep{jazwinski1970stochastic}. Approximating all instances of covariance and mean by Monte-Carlo estimates results in the ensemble update equation, see, e.g.~\cite{Evensen2022}
\begin{equation}
    M^{a} = M^{p} + C_{M,g(M)}\left(C_{g(M)} + C_{d}\right)^{-1}(D - g(M^{p}) + E).
\label{eq:analisys}
\end{equation}
Here, $M$ denotes the ensemble matrix where each column is a realization of the earth model: 
$M^p$ is the prior ensemble for a given step; 
$M^a$ is the ensemble "analysis" matrix, conditioned to the new measurements. 
The shorthand $g(M)$ denotes that the map from the model to synthetic EM-log has been applied to all columns (ensemble members) of $M$, $D$ is an ensemble matrix where each column is a copy of the current observation, and $E$ is an ensemble matrix of measurement perturbations where each column ($\epsilon_j$) is a realization of the Gaussian measurement error $\epsilon_j \propto N(0, C_d)$. $C_d$ denotes the covariance matrix for the current measurements, $C_{M,g(M)}$ denotes the Monte-Carlo estimate of the cross-covariance between the model and the predicted data, and $C_{g(M)}$ denotes the Monte-Carlo estimate of the auto-covariance matrix for the predicted data.

The updated "analysis" ensemble of models, $M^a$, from the EnKF is the input to the optimization.

\subsection{Optimization with dynamic programming for decisions}

Our workflow incorporates an optimization step using  dynamic programming to optimize the remaining drilling trajectory based on continously updated GAN-geomodels. 
This method effectively navigates the high-dimensional space of possible trajectories, identifying the path that maximizes the operational objective by balancing geological target zones and operational constraints.

For each realization of GAN-geomodel in the "analysis" ensemble, our DP method constructs a reward matrix \(R\) that quantifies the geological and operational desirability of drilling various segments.
Each entry \( R_{i,j}(m) \) in this matrix represents the potential reward of steering the drill bit from point \( i \) to point \( j \), given a realization of GAN-geomodel $m$. 
The goal is to find a sequential path $\pi^*_k(m)$ among the potential paths $\pi_k$  (starting from point $k$) that maximizes the cumulative reward:
\todoRev{3}{RESOLVED! Please explain in more detail this reward function.}
\begin{align}
\pi^*_k(m)
&=
\arg\max_{\pi_k}
\sum_{(i,j)\in\pi_k} R_{i,j}(m),
\label{eq:dp}
\\
R_{i,j}(m)
&=
r_{i,j}(m)
+
\max_{j\to l} R_{j,l}(m),
\end{align}
where, \(r_{i,j}(m)\) is the immediate reward (based on the selected reward function) for choosing the segment \((i \!\to\! j)\) under realization \(m\).
The maximization \(\max_{j\to l}\) is taken over all admissible successors \(l\) of \(j\).  
This dynamic-programming recursion is evaluated in reverse order over the grid so that each entry \(R_{i,j}(m)\) contains both the immediate reward for the step \(i\!\to\! j\) and the optimal continuation reward from \(j\) onward.  
The optimal path \(\pi^*_k(m)\) is recovered by selecting, at each step, the successor \(j\) that maximizes \(R_{i,j}(m)\).
The \(R_{i,j}(m)\)  matrices are computed efficiently using dynamic programming principles \citep{cormen2022introduction} with a single pass through each entry in $R_{i,j}(m)$.

The DP approach guarantees optimal paths within each realization, providing long-term adaptability to the uncertain environment. However, making a decision under uncertainty requires considering all options and committing to a single recommendation.
\linelabel{robustdecisiondp}
\todoRev{3}{RESOLVED: Since the forward model is a NN I assume it is quite fast. So why this naive-optimistic approach needed, and for that
matter why even ADP, why not use traditional DP that guarantees global optimal solutions?}

For that, at each step (starting from point $k$), the robust optimization selects the single next decision $\pi^{*1}_{k}(M^a)$, an admissible segment or stopping based on the maximum average expected reward:
\begin{equation}
\pi^{*1}_{k}(M^a) = \arg\max_{j, (k,j) \in \pi}
\left(
\frac{1}{N_e}
\sum_{n \in N_e}
\left[
R_{k,j}(m_n) + \pi^*_j(m_n)
\right]
\right).
\end{equation}
Since the ensemble of GAN-geomodels is updated after drilling each segment, the matrices $\pi^*_j(m_n)$ need to be recomputed against new ensemble members. 
Skipping global optimization under uncertainty makes 'naive-optimistic' DP computations more efficient \cite{alyaev2019decision}, and the gridded representation of geology in this study allows to remove dynamic-programming approximations used by \citet{alyaev2019decision}. \linelabel{nodpapproximations}
At the same time, the one-step robust optimization accounts for uncertainties. 
\linelabel{robustOpt}
Not spending extra time to reconstruct entire trajectory is sensible since future updates in the model often result in different optimal modes of trajectories following different potential drilling targets. 

The updating and optimization process continues until the target depth is reached or further drilling is unprofitable, ensuring that each step maximizes the immediate reward while accounting for the possible future gain. 
By combining AI-driven geomodeling with advanced optimization techniques, the proposed workflow establishes a comprehensive framework for geosteering, enhancing the efficiency, reproducibility, and robustness of drilling decisions.

\begin{figure}
    \centering
\includegraphics[width=0.8\textwidth]{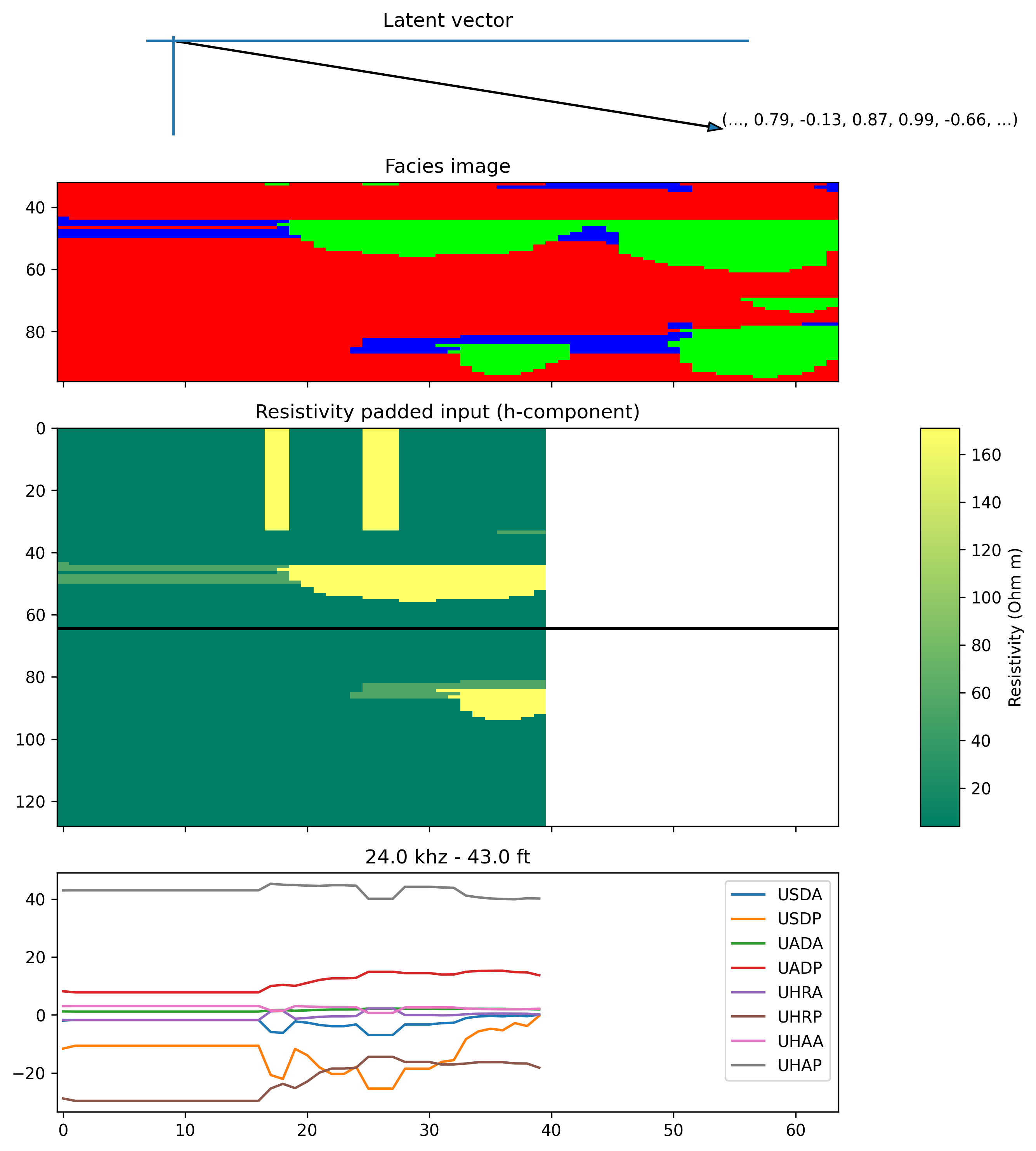}
    \caption{The synthetic truth for the reference case.
    The figure shows the model mapped through all stages of modeling sequence: 
    60-dimensional latent vector, 
    facies image,
    resistivity columns (1 to 40) padded for the FNN (horizontal component visualized),
    logs along the indicated line for a selected transmitter-receiver configuration. 
    The axes show cell indexes of the data.
    Each log point is only sensitive to a single vertical column. The curves' shape indicate sensitivity of measurements to geolgical changes. 
}
    \label{fig:truth}
\end{figure}

\begin{figure}
    \centering
    \includegraphics[width=0.8\textwidth]{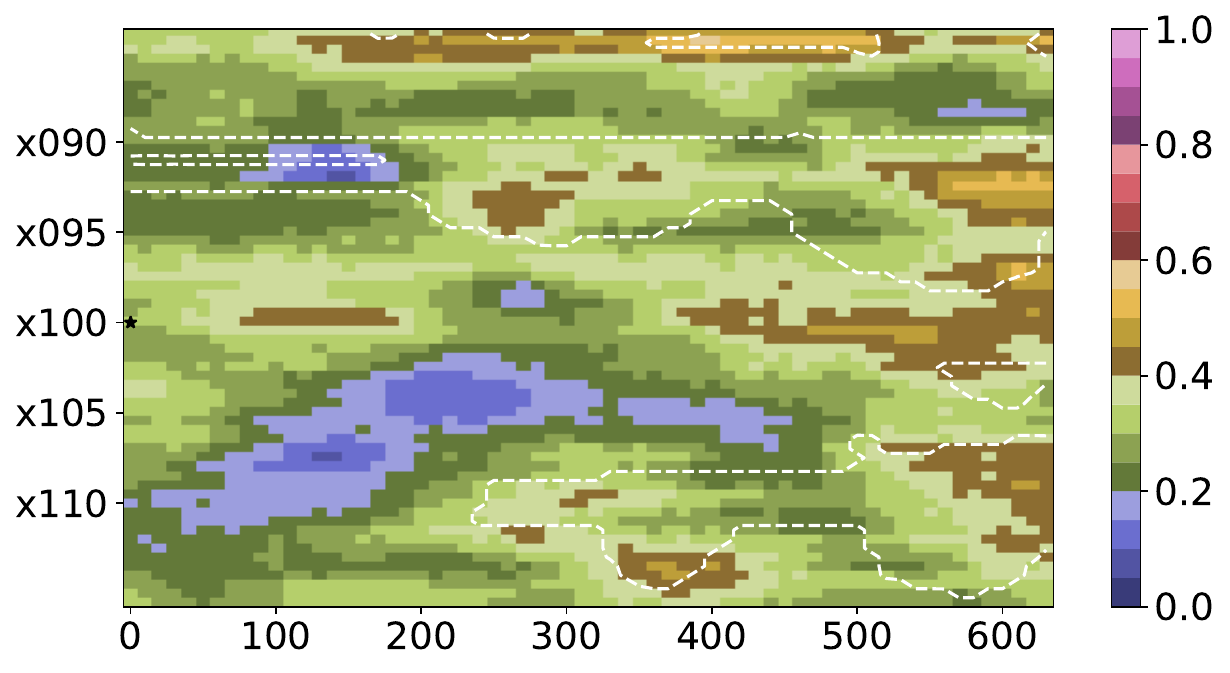}
    \caption{The prior model at the start of the operation with a star indicating the kick-off drilling position. The prior indicates approximately 40\% probability of a sand body in the horizontal direction ahead of the bit.
    The image shows the probability of sand (either channel or crevasse) for every point based on the ensemble average. The white outline indicates the synthetic truth unknown to the DSS.}
    \label{fig:prior}
\end{figure}

\section{Results}
\label{sec:results}
As the reference synthetic test case for the decision-support workflow, we use the same synthetic truth as in \citet{alyaev2021probabilistic}, see Figure \ref{fig:truth}. 
The workflow tests start with the same prior ensemble, with initial drilling location in the middle of the GAN-geomodel.
In Subsection~\ref{sec:biasedCase}, we increase complexity by selecting a synthetic truth away from the prior.

\subsection{Prior ensemble}

The prior consists of 250 model vectors sampled from the same distribution as in \citet{alyaev2021probabilistic}. 
\todoRev{3}{RESOLVED! Why is the prior probability not the same everywhere?}
At the start of drilling, it translates to an ensemble of GAN-geomodels with a large uncertainty in facies distribution, see Figure~\ref{fig:prior}.
Even though the prior vectors are distributed as a multi-Gaussian, their projections to geological facies follows the geological setting. Therefore, despite a high uncertainty in the prior model, some areas have higher probability of sand than others. 
\linelabel{priorprobability}
The drilling starts horizontally in a floodplain shale indicated by the star in Figure~\ref{fig:prior}, but the prior indicates an approximately 40\% probability of a sand body in the horizontal direction ahead of the bit.
Knowing the synthetic truth, based on intuition, we would like the DSS to reduce the uncertainty based on the measurements and redirect the well to the top channel sequence at a depth of x090, with the bottom channel sequence at a depth of x110 being a worse option.

\subsection{Objective function and constraints}

\begin{figure}
    \centering
    \includegraphics[width=0.8\textwidth]{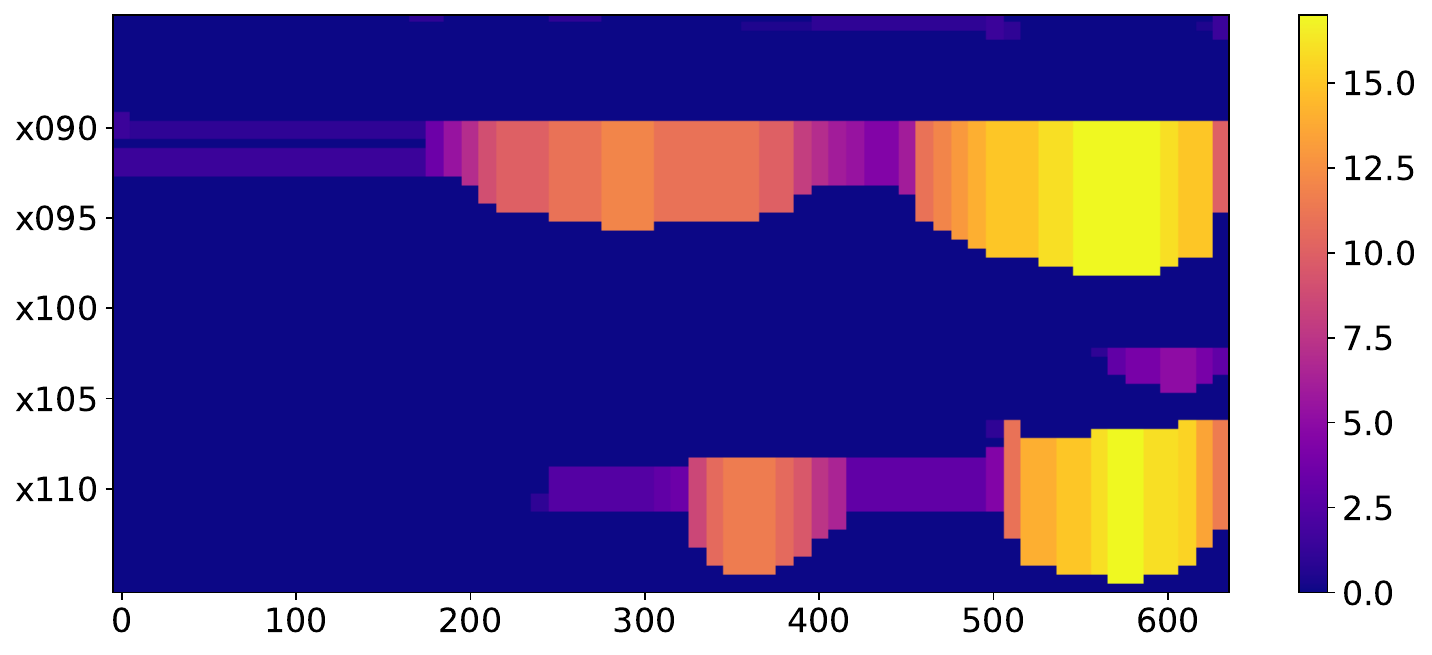}
    \caption{The weighted reward image for the synthetic truth, showing the relative value for reaching each of the cells. The cell values are computed according to Equation (\ref{eq:reward}).}
    \label{fig:rewardCj}
\end{figure}

In a systematic decision analytics framework \citep{kullawan2014decision}, assigning appropriate weights to different geological targets and estimating relative drilling costs are crucial for generating accurate reward matrices. 
The weights formalize the relative importance or desirability of drilling through different geological features. In this synthetic study, we use the following objective function:
\begin{equation}
    r_{i,j} = - r_{i,j}^{\textrm{drill}} + r_{j}^{\textrm{reser}},
\end{equation}
where $r_{i,j}^{\textrm{drill}}$ is the drilling cost proportional to the drilled distance between point $i$, and $j$ (set to $0.02$ per meter drilled) and $r_j$ is the reward for reaching the target at cell $j$. It is defined as:
\begin{equation}
    r_j^{\textrm{reser}} = \int_{k=j_0}^{j_l} w(f_k) dz,
    \label{eq:reward}
\end{equation}
where $j_0$ and $j_l$ are limits of continuous sand (crevasse or channel) in the vertical direction, and $w(f_k)$ is the weight based on the facies type in each cell $f_k$.
\todoRev{2}{RESOLVED! Does the weighting function in Eq 7 cover all options? Or is a weight of zero assigned to non-saturated values? If so it should be
included as a third case as indicated in Table 1.}
We define weights $w(f_k)$ for key geological targets based on their facies type. For this study, the weights are defined as follows:
\begin{equation}
w(f_k)  = 
\begin{cases}
1.0 & \text{if Oil-Saturated Sand Channel},\\
0.5 & \text{if Partially-Saturated Crevasse Splay Sand},\\
0.0 & \text{if Background Shale (elsewhere)}.
\end{cases}
\label{eq:rewardCases}
\end{equation}
The reward image composed of $r_{j}^{\textrm{reser}}$-values for the synthetic true model, unknown to the DSS, is shown in Figure~\ref{fig:rewardCj}. While the reward function is relatively simple in this simplified geological setting, the complexity of the global optimization arises from the possibility of selecting from several alternative target sand bodies in a low net-to-gross setting. This setup creates a non-trivial multi-target optimization problem \cite{alyaev2021interactive}, additionally enriched with lateral geological complexity.
\linelabel{optimizationComplexityTable1}
\todoRev{2}{RESOLVED: Table 1: The facies used here have a fairly big contrast in terms of oil content, which may make the optimization fairly constrained.
Worth commenting on?}

The DP algorithm selects a path where each consecutive segment can go horizontally or diagonally, one cell up or down. 
The weighted reward images constructed for GAN-geomodel realizations updated at each step define the desirability of various paths. They inform the DP algorithm's selection of the optimal drilling trajectory, taking into account both geological and operational factors. 

\subsection{Data-assimilation setup}
\todoRev{3}{RESOLVED!. How was this 10\% chosen? It will be valuable to understand how the magnitude of noise impacts the selection of
the optimal trajectory.}
We perform data assimilation by applying the analysis step (Equation (\ref{eq:analisys})) from the Ensemble Kalman filter. 
In practice, we utilize ensemble smoother method (equivalent to EnKF in this scenario) implemented in the open-source PET library \citep{fossum2024ensemble}.
The sensitivity of our AI-based EM model closely resembles the UDAR tool \citep{alyaev2021modeling}; the log responses to lateral geological changes are visualized for the true model in Figure~\ref{fig:truth}.

We simulate a challenging scenario in which we apply 10\% relative noise to all 56 components of the EM tool. 
\linelabel{noiselevel}
This higher-than-realistic noise level
keeps the uncertainty relatively high for the simplified facies model with high contrasts.
Consequently the setup makes a challenging case for  optimization, emphasizing the workflow's core capability: robust decision recommendations through generative-AI uncertainty quantification, even in low-data-quality scenarios.
\linelabel{challengingUncertainty}

\begin{figure}
\centering
\includegraphics[width=0.9\textwidth]{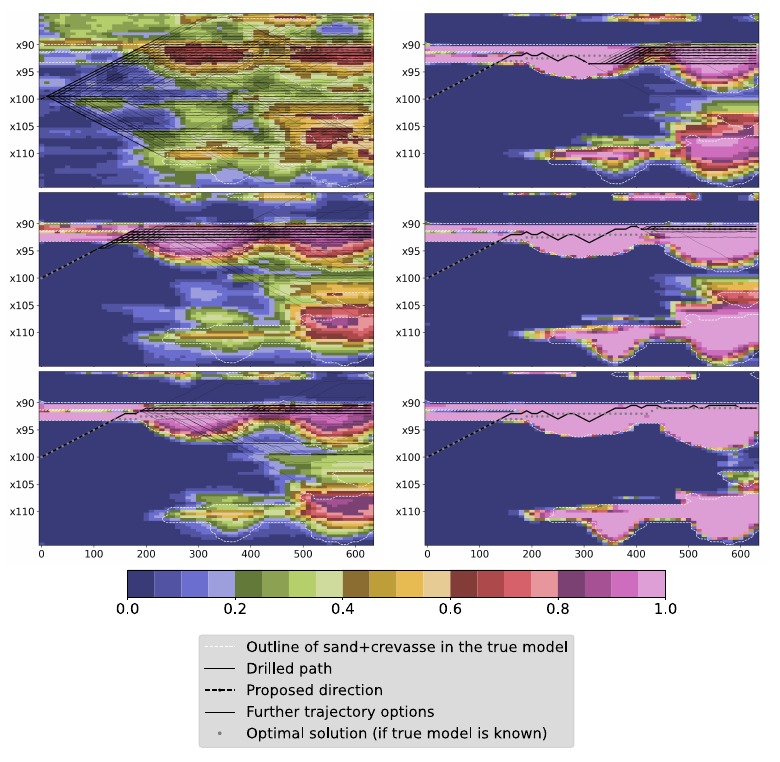}
   \caption{Illustration of the proposed decision-support (DS) workflow applied to a synthetic scenario. This figure details the step-by-step decision recommendations provided by the DS, indicated by the proposed-direction dashed line and advancing the bit in the next sub-figures. The "fan" of thin dashed trajectories, which start from the recommended decision, indicates path possibilities optimal for individual realizations in the ensemble.
    The trajectory paths in the fan are perturbed: thicker line agglomerates correspond to more trajectories targeting each area.
    The scenario begins with an initial geomodel uncertainty, which gradually decreases as EM measurements are collected during drilling along the drilled path. The background images show the probability of sand (channel or crevasse) after data assimilation at the current step. 
    The outline of the true model and the optimal solution within the true setup are provided as a reference but are unknown to the DS.}
    \label{fig:results}
\end{figure}

\subsection{The reference case results}
\label{sec:refResults}

\todoRev{2}{For the latter case, demonstrating a single model/realization of the algorithm in 2D is not sufficient for a journal paper. Additional examples are needed. Any new examples should both consider different truth cases as well as different types of facies structures for the prior.}

\todoRev{2}{Section 3.4 is quite verbose and describes what the algorithm ends up doing for the reference case, but fails to assess how the algorithm
is distinguishing itself relative to other approaches. For instance, if I look at the initial estimation in the upper left of Figure 5 the yellow
regions are both the closest (to the initial position) and largest volumes. It is not obvious to me that the trajectory would be much
different if a greedy strategy based on the initial estimation would be employed. I think the problem would be much more interesting in
3D, as path optimization in 2D as shown in this paper is already quite constrained.}

Figure~\ref{fig:results} illustrates the application of the DS workflow to the reference case step by step. Here we try to justify the observed behaviour of the system.

The "landing" phase (Figure~\ref{fig:results}, column 1) starts from the data assimilation step one, which reveals the misfit of the initial prior (Figure~\ref{fig:prior}) with the true model. After the data assimilation, the estimated probability of sand around the bit becomes low, indicating drilling in the shale. 

The DS optimization finds a suitable long-term target within each of the realizations. The resulting trajectory "fan" for Step 1 spans most of the model because the uncertainty is still high. Due to the GAN-geomodel structure, each of the realizations is geologically plausible, but the outline of the geo-bodies in less probable realization groups are averaged out in the figures. Note that we perturb the trajectory paths so that thick agglomerations of lines highlight the areas targeted across many realizations. At Step 1, the DSS recommendation is to build up the angle.

By Step 11, shown in the second row of Figure \ref{fig:results}, as the EM tool nears the crevasse boundary, the assimilated data clearly outlines a reservoir layer above the bit, predicting the channel-body sequence horizontally due to geological continuity learned by the GAN. 
At Step 11, most of the long-term targets point at the preferred top reservoir sequence, and the DSS recommendation is to hold the angle and continue drilling up. 
We note that in this situation, a most straightforward 'myopic' implementation \cite{alyaev2018you} of a greedy algorithm will recommend terminating the operation, since the immediate reward is zero and is less than the cost of drilling further. \linelabel{greedy}

When the measurement tool reaches the sand at Step 19 (row 3, column 1 in Figure \ref{fig:results}), the well 'lands' in the first sand body in the x090 sequence.
The measurements start detecting the sand bodies in the top of the model.
At this step, the decision is to drop and drill horizontally along the sand-layer roof.

In column 2 of Figure \ref{fig:results}, we observe the recommended well trajectory as the DS follows the x090 sand sequence. 
Simultaneously, the workflow maps the sand bodies around the drilling trajectory. 
Redundant UDAR measurements, combined with GAN-captured geological correlations, allow the capture of sand-body outlines throughout the model despite significant noise in the logs.

At Step 31 (row 1, column 2 in Figure \ref{fig:results}), as the uncertainty of the sand-body shape reduces the drilling is directed towards the sand-layer roof.
At Step 41 (row 2, column 2 in Figure \ref{fig:results}), the bit reaches the crevasse splay between two large sand bodies in the sequence. The primary DS target at this point is continuing in the x090 sand.

At Step 60 (row 3, column 2 in Figure \ref{fig:results}), we see the final part of the reference-case operation. The 60 assimilated log locations have significantly reduced the uncertainty and mapped out the drilled and the surrounding sands. 
Moreover, through the correlations learned by the GAN, the information also extends further away from the well, highlighting a channel sequence continuation towards the right of the model. The sequence at the top of the model is well-identified. The channel stacking below the bit location at Step 60 is picked up, but the shapes cannot be correctly identified. 

Despite very low uncertainty in the end of the synthetic operation, there are some artifacts along the top and the bottom sands. 
The probability of the artifacts' location (especially at the top of the model) increases as drilling progresses.
The reason is the global correlations and that the EnKF does not reuse the previous data, resulting in possible "forgetting" and inflating erroneous correlations when the problem is not well-posed. On the other hand, the global correlations help identify the sand-layer stacking at x105 unseen by the measurements. 

\todoRev{3}{RESOLVED!: It will be very useful to identify the optimal drilling trajectory if the true facies is known exactly, so we can compare
the true optimal trajectory to these results. Additionally it will be good to compare the results of "naive-optimistic" algorithm with
traditional ADP.}

\subsection{Comparison of different logging tools within the workflow}
\label{sec:comparison}

This paper introduces a fully open-source modeling sequence with the modern UDAR tool configuration. In this section, we compare it to the setup from the original conference paper \cite{alyaev2024distinguish}. 
The forward EM-model setup in \cite{alyaev2024distinguish} utilized the previous-generation "extra-deep" EM tool, which had 13 curves compared to 56 in the new setup. The older tool was also limited to detecting only 3 contrasts above and below the measurement location.

\begin{figure}
\centering
\includegraphics[width=0.9\textwidth]{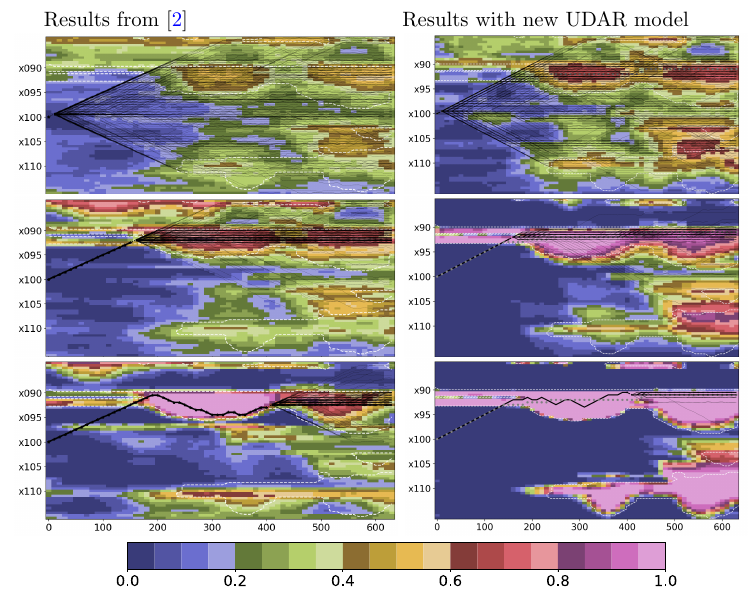}
\caption{
    Comparison of the uncertainty quantification and decisions for the workflow equipped with two different logging tools.
    The left column shows results with extra-deep EM logs (from \citet{alyaev2024distinguish}), while the right column shows results with UDAR tool (same as in Figure \ref{fig:results}).}
    \label{fig:comparison}
\end{figure}

Figure \ref{fig:comparison} compares the two versions of the proposed workflow. The higher sensitivity of the UDAR tool enables mapping not only the layer where drilling occurs, but also the other layers throughout the model. 
On the positive side, lower uncertainty results in a trajectory that is closer to the true optimal (gray dots). 
At the same time, low uncertainty in the UDAR-based case indicates potential ensemble collapse: when the ensemble models are nearly identical, and Bayesian gradients are not well-defined. 
In several tests (outside the scope of this paper), where the true model is selected further away from the prior distribution of the model vectors, we indeed observed failures during data assimilation, resulting in low variance but poor data match; see Section \ref{sec:code} for the reproducible code.
To mitigate such issues, one can select larger ensemble sizes or try an alternative data-assimilation method, such as the particle filter \cite{muhammad2025high}.

\subsection{Case with biased initial ensemble}
\label{sec:biasedCase}

For this test case, we use a synthetic truth generated by flipping several components (45 to 59) of the model vector used to generate the synthetic true model. 
Thus, the true model is shifted away from the ensemble mean in the model space, making it more challenging for the EnKF to capture the correct uncertainty picture in this scenario.

\begin{figure}
\centering
\includegraphics[width=0.9\textwidth]{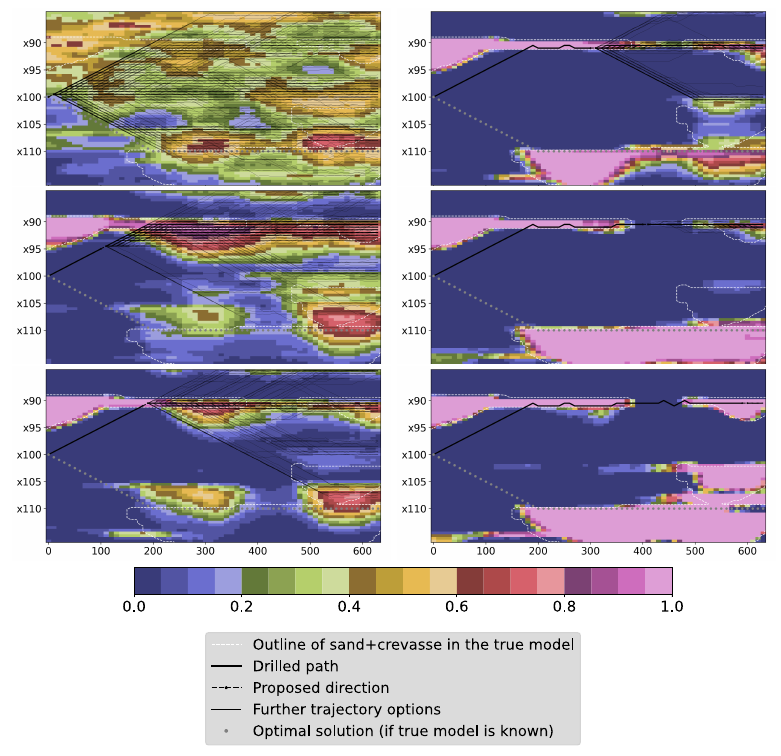}
        \caption{
    A workflow applied to a case with biased ensemble.
    The drilling sequence following decision-support (DS) workflow uses the same prior ensemble as in Figure~\ref{fig:results}, but the synthetic truth is shifted away from the ensemble mean in the model space.
    The initial geomodel uncertainty, which gradually decreases and shifts as EM measurements are collected during drilling along the drilled path. 
    The background images show the probability of sand (channel or crevasse) after data assimilation at the current step. 
    The outline of the biased true model and the optimal solution within it are provided as a reference but are unknown to the DS.}
    \label{fig:resultsBiased}
\end{figure}

As in the reference case from the previous sections, the measurements indicate a sand-body above the starting point of the operation, see Figure \ref{fig:resultsBiased} (row 1, column 1). 
The higher probability of oil-saturated sands above the starting location in most of the updated ensemble, results in the recommendation to steer up. 
Unlike the previous case, the true model has much larger sand layers at around x110, which would give the optimal well placement.

The assimilated measurements by step 11 (row 2, column 1 in Figure \ref{fig:resultsBiased}) reduce uncertainty in the prior. While the shape of the sand layer around x90 is estimated from the measurements, the continuation of the sequence and sand-channel stacking follows the structures typical to the prior. 
These structures resemble more the outline of the true model for the reference case from Figure \ref{fig:results}. This yields the same steering recommendations as in the previous case.

At step 19 (row 3, column 1 in Figure \ref{fig:resultsBiased}), there is already sensitivity to the bottom layer. 
However, a large discrepancy between the prior ensemble and the observations does not allow for the recovery of the true location of the x110 sand sequence. Same can be observed in the second column in Figure \ref{fig:resultsBiased}: the sand bodies are not predicted until the observations from drilling above these sands are obtained.
In this biased case, the final model contains more artifacts from erroneous correlations. 
As the measurements do not identify oil-saturated sands in x110 until late in the operation, the DS consistently (with itself) stays within the initially targeted layer.

In summary, the application of the gen-AI DS workflow to the synthetic cases effectively demonstrates the system's capability to adaptively refine its drilling strategy based on real-time data assimilation and the GAN-learned geological knowledge. 
The iterative updates reduce uncertainty and enhance the accuracy of the geological interpretation, ultimately leading to informed decision-making throughout the drilling process, consistent with uncertainty quantification. 
Despite obvious room for further improvement, the results underscore the potential of an integrated system combining advanced AI techniques with data assimilation to optimize well placement and improve reservoir mapping. \linelabel{clearnoimprovement}
\todoRev{2}{RESOLVED! Together with later statements about further improvements in Section 3.4 gives the impression that this work is incremental and will soon be followed by additional papers.}

\section{Conclusions}
\label{sec:concl}

We have introduced a complete GAN-enabled decision support workflow for assisted geosteering in operations with high geological complexity.
The positive results observed in the presented synthetic reference case demonstrate the effectiveness of the workflow in refining drilling strategies through adaptive real-time data assimilation and AI-driven geological modeling. 
Additional tests demonstrate the workflow's robustness under different levels of uncertainty and varying quality of priors.
These results validate the potential of our approach to improve decision-making processes in geosteering operations.

By sharing the workflow implementation as an open-source code (see Section~\ref{sec:code}), we aim to establish a foundation for the emerging field of real-time drilling supported by generative AI. 
Such an operational, quantitative, decision-driven approach holds great promise for transforming traditional geoscience practices and their operational applications.

Despite advancements enabled by generative-AI geomodeling combined with ensemble data assimilation, there is significant potential for future work, starting with addressing the artifacts observed in the model. 
Possible solutions are improved data assimilation techniques and new generative-AI geomodeling. 
The next generation of gen-AI will also unlock applications for 3D geosteering.
\linelabel{futureWork}
Adapting the workflow to field operations, including those with less data, will also require incorporating more realistic decision objectives and constraints and developing better decision-making methods. 
These efforts will enhance the robustness and applicability of the proposed workflow, driving innovation in geosteering and geoscience in general.

\section{Supporting Data and Code Availability}
\label{sec:code}
The code produced for this paper and running instructions are available in a public GitHub repository. 
You can access the repository at the following link: \url{https://github.com/geosteering-no/DISTINGUISH-WF}. 
\linelabel{newcode}
The dependencies are automatically downloaded from the "geosteering-no" portal: \url{https://github.com/geosteering-no}.
The pre-trained models are distributed through git Large-File Storage (LFS) \url{https://gitlab.norceresearch.no/saly/image_to_log_weights}.

\section{CRediT author statement}
\textbf{SA}: Conceptualization; Methodology; Software; Validation; Formal analysis; Writing - Original Draft; Writing - Review and Editing; Visualization; Supervision; Project administration; Funding acquisition.
\textbf{KF}: Conceptualization; Methodology; Software; Validation; Formal analysis; Investigation; Writing - Original Draft; Writing – Review and Editing; Visualization; Supervision.
\textbf{HJ}: 
Software; Validation; Investigation; Writing - Original Draft; Writing – Review and Editing; Visualization.
\textbf{JT}: Conceptualization; Methodology; Data Curation; Writing - Original Draft; Writing - Review and Editing; Visualization.
\textbf{AE}: Conceptualization; Writing - Review and Editing.

\section{Statement on AI-generated text}
The authors used Open AI ChatGPT to improve draft paragraphs. The AI-generated text was carefully edited to reflect the authors' opinions and perceptions. Most of the text was also processed using Grammarly and its AI Paraphrase Tool to improve formulations and grammar. The authors take full ownership of the study's content and conclusions.

\section{Acknowledgements}

This work is part of the project DISTINGUISH: Decision support using neural networks to predict geological uncertainties when geosteering, funded by Aker BP, Equinor, and the Research Council of Norway (RCN PETROMAKS2 project no. 344236).
We gratefully acknowledge Aspentech for providing an academic software license for Aspen RMS\textsuperscript{TM} used for creating the geomodel dataset. 
\bibliographystyle{plainnat}
\bibliography{lib}

\section{Appendix}

\subsection{Geological setting and the training dataset}

The present study employs a training dataset of 2D sections extracted from a synthetic 3D geological model generated using standard industrial reservoir modeling software. The same geological model was employed by \citet{alyaev2021probabilistic}, but in this paper we take advantage of the new EM simulator and include anisotropy in the shale layers.

\begin{figure}
    \centering
    \includegraphics[width=0.8\textwidth]{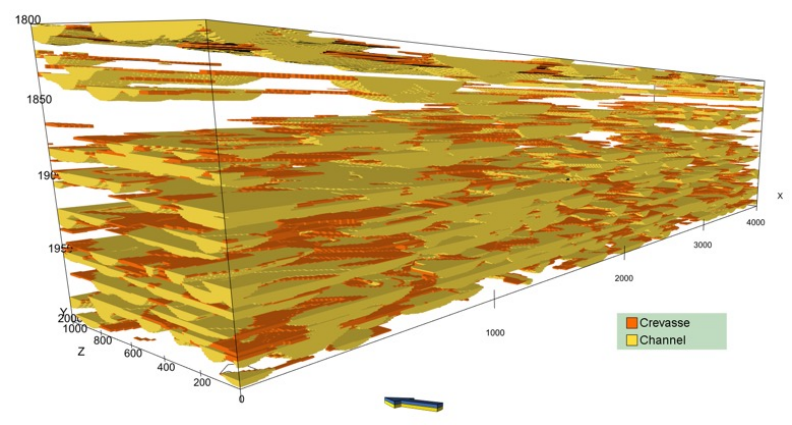}
    \caption{The facies structure of the synthetic 3D geomodel used for training. The background floodplain shale facies are filtered out to highlight the distribution of channels and surrounding crevasse splays, underscoring the complexity of the model's 3D depositional architecture. Note 5x vertical exaggeration.
}
    \label{fig:geomodel3d}
\end{figure}

The model consists of an un-faulted, orthogonal grid measuring 4000m x 1000m x 200m (XYZ), with a resolution of 10m x 10m x 0.5m, yielding 16 million grid cells. The depth range of the model is set at 1800m to 2000m TVD. The depositional architecture in the model is based on geometric data from outcrops of the fluvial, low net/gross, lower Williams Fork Formation (Cretaceous Mesa Verde Group) at Coal Canyon, Colorado, USA \citet{pranter2011static,pranter2014fluvial,trampush2017identifying}, comprising channel-, crevasse-splay-, and overbank/floodplain shaly facies, see Figure \ref{fig:geomodel3d}. In the model, the flow direction of the fluvial system is set to $45 \pm 10$ degrees, with no trend functions constraining the spatial distribution of channels. For detailed parameter settings used in the facies model setup, see \citet{alyaev2021probabilistic}. 
Petrophysical values used here are simplified and assumed constant, albeit anisotropic for each facies; see Table~\ref{tab:facies}. 
The table also shows a simple constant reward for drilling in each facies type, correlated to porosity and expected oil content within the facies.

\begin{table}[h!]
\centering
\begin{tabular}{|l|r|c|r|r|r|}
\hline
\textbf{Facies} & \textbf{Porosity (\%)} & \textbf{Oil Content} 
& 
\textbf{$R_h$, (ohm-m)} 
&
\textbf{$R_v$, (ohm-m)} 
& \textbf{Reward} \\
\hline
Channel Sand & 20 & Oil-saturated & 171.0 & 171.0 & 1.0 \\
Crevasse Splay Sand & 12 & Partially & 55.0 & 85.0 & 0.5 \\
Floodplain Shale & 2 & No & 4.0 & 12.0 & 0.0 \\
\hline
\end{tabular}
\caption{Mean porosity, oil content, resistivity, and "per-pixel" reward $C$ for different facies types.}
\label{tab:facies}
\end{table}

The training dataset consists of 2D XZ-sections measuring 64 x 64 cells (i.e., 640m x 32m) sampled throughout the geo-model. The third dimension of the sampled sections represents a facies probability index. 

\subsection{Training models in the mapping sequence}
\label{sec:appendixTraining}

This appendix gives the details on training procedures for parts of the mapping sequence which are considered pre-trained at the workflow application stage. 
The components are the Generative Adversarial Network GAN for generating geological realizations and a Forward deep Neural Network (FNN) proxy that converts local geological configurations to simulated Ultra-Deep Azimuthal Resistivity (UDAR) measurements. Despite the traditional resistivity name, UDAR are phase and attenuation  electromagnetic measurements influenced by subsurface resistivity, but not directly measuring it. 

\subsubsection{Generative Adversarial Network (GAN)}

The GAN used in the workflow follows the setup first introduced by
\citet{alyaev2021probabilistic}.  
It consists of a deep convolutional generator that maps
60-dimensional latent vectors to 64\,$\times$\,64 facies images with three channels (channel, crevasse, and floodplain shale), and a discriminator (critic) trained to distinguish real geomodel slices from generated ones.  
The latent prior, sampled from an independent, factorized Gaussian $\mathcal{N}(0, I_{60})$ typical of GANs, provides a simple, uncorrelated input space for the generator.
\linelabel{gaussianity}
Both networks are trained simultaneously by "playing" a min-max game using the Wasserstein loss \citep{arjovsky2017wassersteingan}, thereby providing smooth latent–space mappings \cite{elsheikh_gan_2026_inpress} and stable convergence, both of which are essential for ensemble updates in data assimilation.

\todoRev{3}{\ans{gaussianity}
Are the 60 variables considered independent? If yes, during the offline training, how is this assumption honored? Is the assumption of Guassianity valid? If so, why?}

\begin{table}[h!]
\centering
\caption{Key GAN configuration used in training.}
\label{tab:gan_hyper}
\begin{tabular}{l l}
\hline
\textbf{Parameter} & \textbf{Value / Description} \\ \hline
Generator input dimension & 60 \\
Output image size & 64\,$\times$\,64\,$\times$\,3 \\
Architecture type & Deep convolutional (DCGAN-style) \\
Loss function & Wasserstein loss \\
Training data & 2D slices from 3D fluvial geomodel (Williams Fork Formation) \\
Facies classes & Channel, crevasse, and floodplain shale \\
\hline
\end{tabular}
\end{table}

\begin{figure}[h!]
\centering
\includegraphics[width=0.85\textwidth]{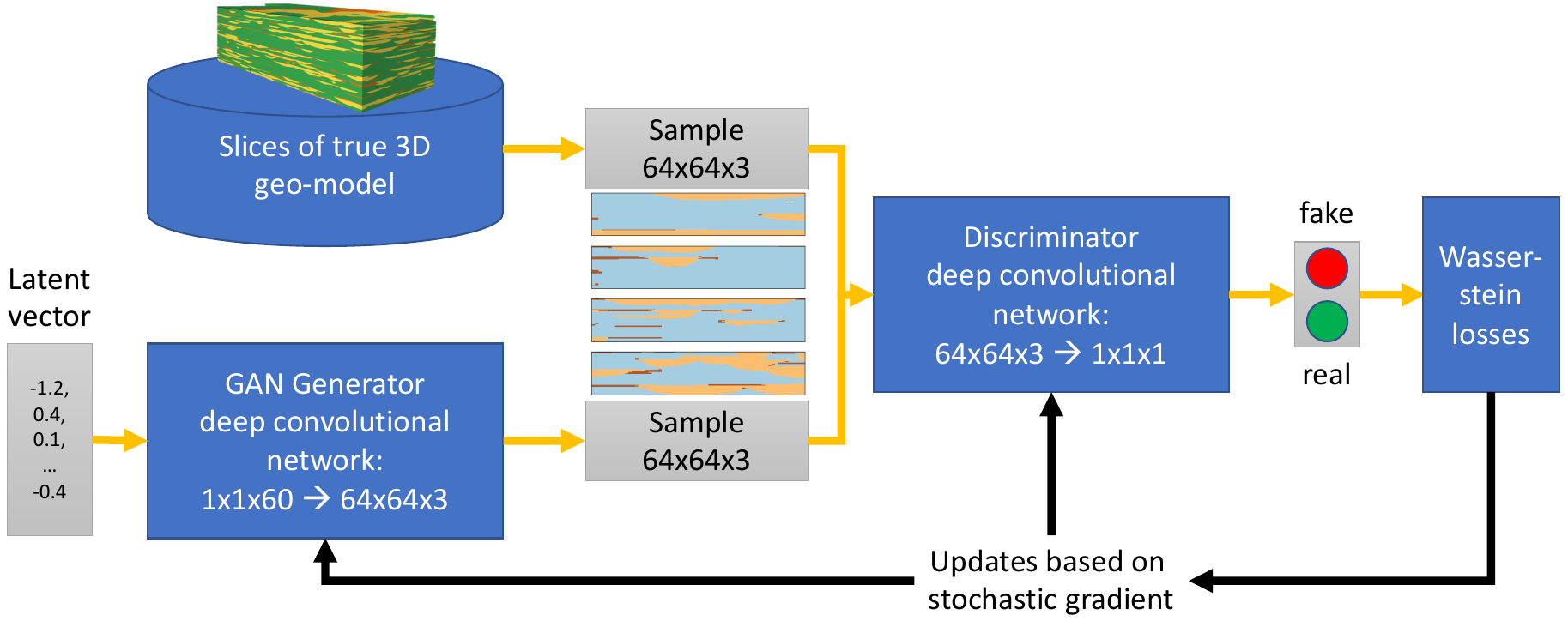}
\caption{Architecture of the GAN used for fluvial facies generation, showing
the generator-discriminator configuration and the use of Wasserstein losses.
Reproduced under the CC-BY 4.0 license from the original 2021 training
materials of \citet{alyaev2021probabilistic}.}
\label{fig:gan_architecture}
\end{figure}

\subsubsection{Forward Neural Network (FNN)}

The forward model for electromagnetic (EM) responses 
is implemented as a
fully convolutional 1-D residual network. 
The model consists of a series of seven nested convolutional blocks, each comprising two 1-D convolution layers with ReLU activation and an internal residual connection, followed by max-pooling that progressively reduces the resolution while increasing the number of feature channels. This produces a deep encoded representation that captures multiscale structure in the input. After the final block, a transposed 1-D convolution expands the compressed sequence back to the desired output length, and a concluding 1×1 convolution projects the feature dimension to the required number of output channels.

The model predicts the real and imaginary parts of the electromagnetic response at the antenna position for the x-, y-, and z-oriented field components, as well as the cross-coupling combinations \emph{xz} and \emph{zx} (1st letter denotes the direction of the transmitter antenna and 2nd letter denotes the direction of the receiver antenna). This gives \textbf{ten} independent components.
In addition it learns and predicts \textbf{eight} commonly-used derived components: Ultra-deep Symmetrized Directional Attenuation and Phase (USDA and USDP), the Ultra-deep Anti-symmetrized Directional Attenuation and Phase (UADA and UADP), the Ultra-deep Harmonic Resistivity Attenuation and Phase (UHRA and UHRP), and finally the Ultra-deep Harmonic Anisotropy Attenuation and Phase (UHAA and UHAP) as described in~\citet{Davydycheva2023}. 
Each of the components is obtained for a tool with two transmitter-receiver pairs (43 and 83 ft apart) operating on a range of frequencies (6, 12, 24, 48, 96 kHz). The \textbf{six} used frequency-distance pairs are:
(6kHz, 83ft), (12kHz, 83ft), (24kHz, 83ft), (24kHz, 43ft), (48kHz, 43ft), (96kHz, 43ft).

In total, the model generates $(10+8)\times6 = 108$ log curves, some of which are not independent. For the workflow testing in this paper we only used the $8\times6=56$ derived components. \linelabel{newtool}

The model is trained using mean-squared error loss and the Adam optimizer for a fixed number of epochs, with periodic checkpointing. During training, the mean-squared error converged to approximately $\mathbf{O}(10^{-6})$, which is consistent with the error reported in \citet{alyaev2021modeling}.

\begin{table}[h!]
\centering
\caption{Summary of FNN architecture and training setup.}
\label{tab:conv_hyper}
\begin{tabular}{l l}
\hline
\textbf{Parameter} & \textbf{Description} \\ \hline
Network type & 1D convolutional residual network \\
Convolutional blocks & 7 nested blocks with channel growth (40, 80, ..., 280) \\
Block structure & Conv1D → ReLU → Conv1D → ReLU → residual add → MaxPool(2) \\
Kernel size & 64 for all convolution layers \\
Upsampling & ConvTranspose1D to match output sequence length \\
Output projection & 1×1 Conv1D mapping to target channels \\
Activation function & ReLU throughout \\
Loss function & Mean squared error (MSE) \\
Optimizer & Adam, learning rate $10^{-4}$ \\
Batch size & 1000 \\
Training epochs & 1000 \\
Training strategy & Fixed-epoch training with periodic checkpointing \\
Normalization & Feature-wise min–max scaling of input and output \\
Training data & 40.000 precomputed pairs of input/output 1D EM field sequences \\
\hline
\end{tabular}
\end{table}

The GAN and FNN models described above form the offline-trained components of
the workflow. Together, they ensure rapid mapping from latent geological
vectors to synthetic EM logs during online data assimilation and decision
optimization.

\end{document}